# Retrain or not retrain? - efficient pruning methods of deep CNN networks

Marcin Pietron
Department of Electrical Engineering
AGH, University of Science and Technology
Cracow, Poland
pietron@agh.edu.pl

Maciej Wielgosz
Department of Electrical Engineering
AGH, University of Science and Technology
Cracow, Poland
wielgosz@agh.edu.pl

## Abstract

Nowadays, convolutional neural networks (CNN) play a major role in image processing tasks like image classification, object detection, semantic segmentation. Very often CNN networks have from several to hundred stacked layers with several megabytes of weights. One of the possible methods to reduce complexity and memory footprint is pruning. Pruning is a process of removing weights which connect neurons from two adjacent layers in the network. The process of finding near optimal solution with specified drop in accuracy can be more sophisticated when DL model has higher number of convolutional layers. In the paper few approaches based on retraining and no retraining are described and compared together.

## 1 Introduction

The convolutional neural networks are the most popular and efficient model used in many AI tasks. They achieve best results in image classification, semantic segmentation, object detection etc. The reduction of memory capacity and complexity can make use of them in real-time applications like self-driving cars, humanoid robots, drones etc. Therefore compression CNN models is a important step in adapting them in embedding systems and hardware accelerators. One of the step to decrease memory footprint is a pruning process. In case of small convolutional network the complexity of this process is much lower than in larger ones. In very deep CNN models which have from several up to few hundreds of convolutional layers the process of finding near global optimum solution which guarantee acceptable drop in accuracy is quite complex task. Genetic/memetic algorithms, reinforcement learning, random hill climbing or simulated annealing are one of the candidates to solve this problem. In paper algorithm based on RMHC and simulated annealing methods is presented. The pruning process can be done by two major methodologies. First one is a pruning a pre-trained network, the second one is pruning using retraining. The first one is much faster. It needs only an inference step run on a test dataset in each stage/iteration of the algorithm, [2]. In case of mode with retraining pruning can be done after every weight update in training process. In paper there are described and compared approaches using both methodologies.

The Squeezenet model was one of the first approach in which compression by reducing the filters size was used. In this approach architectures of alexnet was modified to create less complex model with same accuracy. Later approaches concentrate more on quantization and pruning [2], [6] as a step that enables compression. In [6] authors present approaches for CNN compression including pruning with retraining. The results for older architectures VGG and AlexNet are presented. In paper [8] authors describe reinforcement learning as a method for choosing channels for structural pruning. In article [7] the SNIP algorithm is described. The algorithm computes gradients during retraining and assigns priorities to weights based on gradients values. The pruning is done using knowledge about importance of weights in a training process. In papers [4], [5] compression for other machine learning models are described in NLP tasks. It is shown that using sparse representations especially it is possible to achieve better results than in baseline models. The paper is organized as follows. The section 2 presents methods for pruning pre-trained networks. There is a basic method and its further enhancements using more complex models analysis. The next section 3 is about pruning with retraining on imagenet, CIFAR10 and CIFAR100 datasets and structural pruning. Finally, in 4 and 5 further work and conclusions are described.

## 2  Pruning with no retraining

After process of training neural model we acquire a set of weights for each trainable layer. These weights are not evenly distributed over the range of possible values for a selected data format. Majority of weights are concentrated around 0 or very close to it. Therefore, their impact on the resulting activation values is not significant. Depending on network implementation specifications, storing weights may require a significant amounts of memory. Applying pruning process to remove some weights has a direct impact on lowering storage requirements.

In this section the approaches based on pre-trained networks are presented. The first one is memetic approach which is based on random hill climbing with few extensions. The parameters to the heuristic were added to optimize and speed up the process of finding local optima solutions. Next, additional more sophisticated analysis was incorporated to previous approach to improve obtained results. These methods analyses energies of 2D filters inside layers and class heat maps. In most presented approaches pruning is a function that set of weights with magnitudes below specific threshold are set to zero value.

### 2.1  Incremental pruning based on random hill climbing

The presented approach for fast pruning is based on random hill climbing and simulated annealing local search. In each iteration, it chooses specified number of layers to be pruned. The layers are chosen using probability distribution based on layers' complexities and sensitivities (eq.1, eq.2, line 4). If a layer is more complex and less sensitive than others, it has more probability to be chosen. In each iteration, layers are pruned by the step which can be different and computed independently for each layer (line 7). If drop in accuracy is higher than given threshold reverse pruning is applied (the step can be cancelled or sparsities of different layers are decreased). Fitness function is a weighted sparsity which is overall memory capacity of current pruned model (line 11). Solution is a simple genotype where each layer is represented as a percentage of weights that were pruned for this layer. Algorithm can use as an option simulated annealing strategy which accepts worse solutions (exploration phase) to have possibility to escape from local optima (line 18-22). In this case, in line 21 a next created solution can be worse than previous solution and will be accepted with specified probability which decreases in each iteration. Algorithm has a ranked list of all k-best solution already found (line 14). It helps to overcome algorithm stagnation by giving opportunity to return to good solutions (line 19). Each layer as it was mentioned earlier has sensitivity parameter which measures latest impacts (number of impacts is defined by window size parameter) of this layer to the drop of accuracy of the model (eq.3, line 13). The layer sensitivity is updated after each iteration in which given layer is pruned (line 13). The step size which indicates pruned for a given layer is computed using current sensitivity value of a layer. If sensitivity is less than acceptable drop in accuracy (threshold) algorithm increases step size and vice versa using eq.4, line 24.

$$\text{probability}_i = \text{size}_i \times (\text{threshold} - \text{sensitivity}_i) \qquad (1)$$

$$\text{policy} = \text{categorical}(\text{probability}) \qquad (2)$$

---

**Algorithm 1** Pruning algorithm
---
1: Input: number_of_iterations
2: Input: drop_in_accuracy_threshold
3: for number_of_iterations do
4:     update_policy()
5:     layer = choose_layer_for_pruning(policy)
6:     if (top_1-baseline) < drop_in_accuracy_threshold then
7:         prune_layer_by_step(step$_l$ )
8:     else
9:         reverse_prune_by_step(step$_l$ )
10:    end if
11:    fitness = compute_new_fitness()
12:    top1 = compute_accuracy()
13:    update_layer_sensitivity(layer)
14:    update_ranked_list()
15:    if fitness < best_fitness then
16:        next_solution = current_solution
17:    else
18:        if SA_Probability < threshold then
19:            next_solution = solution_from_ranked_list()
20:        else
21:            next_solution = current_solution
22:        end if
23:    end if
24:    step$_l$ = update_steps(layer)
25: end for

Table 1: Pruning results

| Name | weighted sparsity | T1 |
|---|---|---|
| vgg16 | 35.3% | -0.8% |
| resnet50 | 32.1% | -1% |
| vgg19 | 32.6% | -1% |
| inception_v3 | 18.1 % | -0.8% |

$$\text{sensitivity}_i^t = \sum((\text{baseline\_acc} - \text{pruned\_acc}_t) \div \text{window}) \qquad (3)$$
$$\text{step}_i^t = \text{step}_i^t + k\ \text{step}_i^t \times (\text{threshold} - \text{sensitivity}_i^t) \qquad (4)$$

The presented algorithm can be run in multi-layer mode in which in one iteration more than one layer can be pruned. In tab.1 there are results achieved using algorithm 1 with constant policy by running 150 iterations. It contains weighted sparsities of pruned models and theirs drops from baseline accuracies.

The threshold drop was set to 1.0. The tab.2 presents results using prioritization mode in which largest layers in a given models were chosen for pruning in the first stage of the algorithm till the drop of accuracy is higher than given priority list drop. After that rest of the layers are pruned. We can observe significant improvements in achieved results. Tab.3 shows results when using dynamic policy updates during algorithm.

### 2.2 2D filter and its activation analysis for further pruning improvements

Improvement presented in this subsection does additional analysis that can explain the internal representation of the model and removes more weights with high probability to not decrease its accuracy. First approach is to compute 2D average filters contributions in a final answer of the network. The next one is to analyze filter contribution in a process of recognition specific class. Each class is analyzed separately and average neurons activations are measured. Then in each layer we can extract region of weights that are less important in whole process of recognition using some threshold of importance. In tab. 4 and tab. 5 there are results presented for these two steps performed on last layer in VGG16 after running alg.1. It shows that is possible to do further pruning to improve little bit sparsity without drop in accuracy.

Table 2: Pruning results with specified prioritization

| Name | weighted sparsity | T1 | prioritization list |
|---|---|---|---|
| vgg16 | 67% | -0.9 | layers 14,15,16 |
| resnet50 | 37% | -1.1 | 5 largest layers |
| vgg19 | 65 | -1.0% | layers 17,18,19 |
| inception_v3 25 | 35% | -0.9 | 8 largest layers |

Table 3: Results of pruning with dynamic policy

| Name | weighted sparsity | T1 |
|---|---|---|
| vgg16 | 65% | -1.0 |
| resnet50 | 35% | -1.0 |
| vgg19 | 60% | -0.9 |
| inception_v3 | 24% | -0.9 |

Table 4: VGG16 with 2D filter analysis

| Name | T1 | pruning | with 2D | drop |
|---|---|---|---|---|
| CIFAR10 | 90.76% | 50% | 52% | -1.0% |
| CIFAR100 | 77.6% | 45% | 47% | -1.0% |

Table 5: VGG16 with 2D filter analysis and filter contributions in a classes recognition

| Name | T1 | pruning | all | Drop |
|---|---|---|---|---|
| CIFAR10 | 90.55 % | 50% | 54% | -1.0 |
| CIFAR100 | 77.5 % | 45% | 50% | -1.0 |

# 3 Pruning with retraining

The methods described in previous section have one main drawback, their weight can be fine tuned during the pruning process to boost models accuracy. The training step can improve accuracy of pruned network by learning weights that were not removed before. In this section results of these methods are presented.

## 3.1 Retraining methods

Retraining is recognized as an effective method for regaining performance of the pruned model. However, it is important to pick a right protocol and retraining parameters. We have examined three different schemes of pruning and retraining:

- simple retraining which without masking
- simple retraining with masking,
- adaptive retraining with boosting.

The first two methods apply a simple retraining procedure after each step of pruning. The procedure can be interleaved with masking operation. This operation prevents application of update step every epoch.
It is implemented by zeroing gradient which otherwise would be applied to the pruned weights. It is worth noting that even without masking the pruned weights are mode prone to be pruned again in the next epoch because there are small. Consequently, the masking operation makes the pruning process more stable since a pool of pruned weights is progressively enlarged without change of coefficients. The change of coefficients occurs in the procedure without masking because it happens from time to time that the pruned weights are larger in some other in the model after update operation. The simple method is limited in its effectiveness mostly because it lacks ability to adopt pruning both in terms of layers of the model and the retraining time. Some layers during selected training epochs are more prone to pruning, which is not taken into account in the simple method. Therefore, we have proposed the retraining with boosting procedure which is given by Alg. 2. The proposed approach Alg. 2 relies on a choice of priority list of the layers which is supposed to be set at the very beginning of the process. The rest of the parameters steps decide how many steps are taken before scale is changes.

---

**Algorithm 2** Retraining with boosting

```
 1: Input: scales
 2: Input: steps
 3: Input: step_size
 4: for number_of_epochs do
 5:     layer = choose_high_priority_layer_for_pruning()
 6:     for layers do
 7:         pick_the_next_layer_from_the_priority_list()
 8:         for scales do
 9:             for steps do
10:                 prune()
11:                 validate()
12:                 if performance_drop < threshold0 then
13:                     if skipped_no < threshold1 then
14:                         mark_layer_done_for_this_iteration()
15:                     else
16:                         mark_layer_done_for_all_iteration()
17:                     end if
18:                     mark_layer_done_for_this_iteration()
19:                     exit()
20:                 end if
21:             end for
22:             step_value   step_value=reduction_factor
23:         end for
24:     end for
25: end for
```

This gradually reduces pruning factor. The scale (refer to Alg. 2 ) decides how many times the step is reduced. Once the model is pruned it is validated with a small dataset to check if the performance drop is not to large. If this is the case the process of pruning is stopped for the given layer in this iteration (epoch) and the algorithm goes to the next layer on the priority list. The pruning process may terminate in a regular fashion when all the steps and scale rates are exhausted. In order to speedup the process a layer which was skipped several times due to the performance drop after pruning is marked as permanently skipped. It is worth noting that a number of epochs should be picked properly in order to satisfy the number of the protocol interactions (number of steps and scale changes).

### 3.2 Results of the pruning and retraining experiments on imagenet

There was series of experiments conducted as presented in Tab. 8, 6 and 7. Different parameters were chosen as well as different strategies were tested. In the first a naive approach was explored as a baseline. The results are presented in Tab.6. We can see that equal pruning of all the layers for 0.2 and 0.3 sparsity led to the boost of the model performance. However, more aggressive pruning of 0.7 equal sparsity resulted in a significant decline of the sparsity. The proposed simple method may be useful when treated as a form of regularization and slight increase of the model sparsity.

It is worth noting that progressive pruning which results are presented in Tab. 7 is much more effective. For instance, the experiment with starting point of 0.1 and progress of 0.01 every epoch (see the last row in Tab. 7) allowed to reach equal sparsity of 53 % after 43 epochs with negligible loss of performance. This method despite its benefits is limited in its capacity to reduce sparsity. In the series of experiments which results were not presented due to limited amount space saturates at about 60 % of sparsity.

Table 6: Results of Resnet-50 simple pruning and retraining

| layers pruned | sparsity | masking | Best T1 | t1 err. | Best T5 | epoch | lr (reduction) | batch size |
|---|---|---|---|---|---|---|---|---|
| None | 0 | None | 76.13 | 0 | 92.862 | 103 | 0.1 (30) | 256 |
| all-0.2 | all layers 0.2 | FALSE | 76.83 | 0.7 | 93.15 | 14 | 1.00E-03 | 256 |
| all-0.3 | all layers 0.3 | FALSE | 76.95 | 0.82 | 93.21 | 68 | 1.00E-03 | 256 |
| all-0.7 | all layers 0.7 | TRUE | 59.55 | -16.58 | 83.62 | 26 | 0.1 | 256 |

Table 7: Results of Resnet-50 progressive pruning and retraining

| layers pruned | sparsity | masking | Best T1 | t1 err. | Best T5 | epoch | lr (reduction) | batch size |
|---|---|---|---|---|---|---|---|---|
| None | 0 | None | 76.13 | 0 | 92.862 | 103 | 0.1 (30) | 256 |
| all-0.2 + 0.1*epoch | all layers 0.3 | FALSE | 76.48 | 0.35 | 93.07 | 1 | 1.00E-03 | 256 |
| all-0.1 + 0.1*epoch | all layers 0.2 | TRUE | 76.56 | 0.43 | 93.1 | 1 | 1.00E-03 | 256 |
| all-0.1 + 0.01*epoch | all_layers_0.53 | TRUE | 76.09 | -0.04 | 93 | 43 | 0.01(30) | 256 |

Table 8: Results of Resnet-50 boosted pruning and retraining

| layers pruned | sparsity | masking | Best T1 | t1 err. | Best T5 | epoch | lr (reduction) | batch size |
|---|---|---|---|---|---|---|---|---|
| None | 0 | None | 76.13 | 0 | 92.862 | 103 | 0.1 (30) | 256 |
| steps:12, scales:2, step:0.05 | weighted: 0.37 | TRUE | 75.12 | -1.01 | 92.57 | 47 | 1.00E-03 | 256 |
| steps:12, scales:2, step:0.05 | weighted: 0.42 | TRUE | 75 | -1.13 | 93.09 | 30 | 1.00E-03 | 256 |
| steps:12, scales:2, step:0.05 | weighted: 0.427 | TRUE | 79.88 | 3.75 | 94.96 | 100 | 1.00E-03 | 256 |
| steps:12, scales:2, step:0.05 | weighted: 0.57 | TRUE | 75.35 | -0.78 | 92.59 | 299 | 1.00E-03 | 256 |
| steps:4, scales:2, step:0.05 | weighted: 0.5137 | TRUE | 75.61 | -0.52 | 92.67 | 431 | 1.00E-03 | 256 |
| steps:6, scales:2, step:0.05 | global: 0.57 | TRUE | 75.52 | -0.61 | 92.68 | 279 | 1.00E-03 | 256 |
| steps:10, scales:2, step:0.05 | global: 0.44 | TRUE | 76.23 | 0.1 | 93 | 141 | 1.00E-03 | 256 |
| steps 2, scale:4, step:0.2 | global: 0.648 | TRUE | 74.51 | -1.62 | 92.19 | 88 | 1.00E-03 | 256 |

The most advanced approach of pruning and retraining in the boosting method given by Alg. 2. Its results are presented in Tab. 8. We can in Tab. 8 that different values of steps and scales lead to huge discrepancies in the results in terms of sparsity. The highest sparsity of 64.8 % was achieved for steps:2, scale:4 and step value: 0.2. This was achieved at the expanse of noticeable loss of the performance. On the other hand small step value, large number of steps and training epochs lead to much lower performance degradation as proved by the experiment with steps:6, scales:2, step value: 0.05 and 279 epochs of training. However, such large number of epochs required approx. 10 days of training time on 8 Nvidia GTX 1080 GPUs. Choice of a proper number of steps, scales and step values should be done individually for each model and ideally facilitated with an optimization algorithm.

During a pruning and retraining operation of a pretrained model with high learning rate, there is a huge degradation of the performance (t1 and t5) in the very first epoch as presented in Fig. 1. In the next epochs the model regains it original performance quite fast. The presented in Fig. 1 resembles in terms of a training pattern most of the experiments showed in Tab. 8.

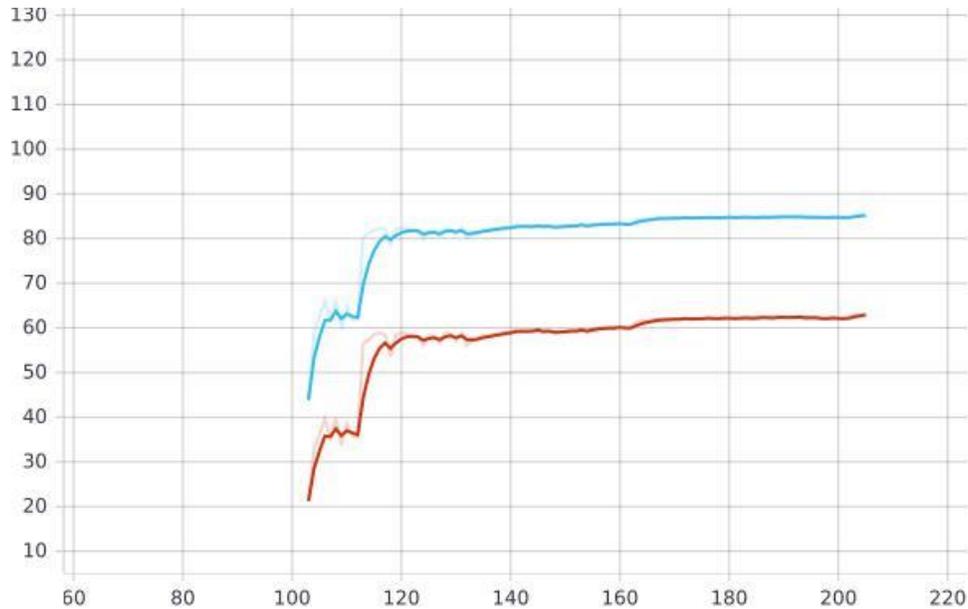

Figure 1: Retraining of the pretrained Resnet50 with global sparsity of 20 %. Retraining starts at 104 epoch. Top5 is marked in blue and Top1 in red.

### 3.3 Pruning with retraining on CIFAR datasets

The similar approach as described in previous section was performed on a CIFAR10 and CIFAR100 datasets. The main difference is that in each step the weights for pruning in each step were chosen using it gradient values. This information gives feedback how important the weight was in former training step (alg. 3). If its significance is less than threshold it is more safe for removing. The results in tab. 9, tab. 10 presents results obtained using algorithm 3. They show significant improvement in obtained sparsity when compare to fast pruning approach.

**Algorithm 3** Pruning algorithm with retraining

```
1:  Input: number_of_epochs
2:  Input: drop_in_accuracy_threshold
3:  Input: init_sparsity
4:  Input: init_step
5:  for number_of_epochs do
6:      layer = choose_layer_for_pruning(policy)
7:      analyze_gradients_and_update_statistics()
8:      if (top_1-baseline) < drop_in_accuracy_threshold then
9:          prune_layer_by_step(stepl)
10:     else
11:         reverse_prune_by_step(stepl)
12:     end if
13:     top1 = compute_accuracy()
14:     update_layer_sensitivity(layer)
15:     stepl = update_layer_step(layer)
16:     masking()
17:     retrain()
18: end for
```

Table 9: Results of fine-grain pruning with retraining (CIFAR10)

| Name | baseline T1 | pruned T1 | pruned size |
|------|-------------|-----------|-------------|
| vgg19 | 92.37 | 91.81 | 2% |
| resnet50 | 95.26 | 94.99 | 8% |

Table 10: Results of fine-grain pruning with retraining (CIFAR100)

| Name | baseline T1 | pruned T1 | pruned size |
|------|-------------|-----------|-------------|
| vgg19 | 70.62 | 70.12 | 5% |
| resnet50 | 78.21 | 77.56 | 22% |

### 3.4 Structural pruning

Structural pruning is a process where blocks of weights are removed. One of the most popular is reducing number of channels in a filter. Using this approach straightforward implementation on many hardware accelerators can speedup original network without any software modification. Reducing the number of channels (chunk of weights) in a pre-trained network usually affects significantly models accuracy. This approach should be mixed with a training steps to minimize the accuracy drop. In presented approach the channels with lowest L1 norm and lowest variance among 2D filters inside given channel were chosen to be removed. The subset of such channels was extracted in each iteration. Then retraining process was started to increase accuracy. The process till accuracy after training step was below threshold given as an input parameter (1%). The results are presented in tab.11, tab.12. It is worth noting that results achieved using this approach are significantly worse than in fine grain pruning.

## 4 Conclusions

The results presented in this paper show quite high disparities in sparsities between pruning with retraining or without retraining. Retraining can significantly improve the drop of accuracy after pruning. During retraining process other aspects like masking, step size of the pruning in current stage of pruning process are very important to achieve better results. The same effect we can observe in fast pruning on pre-trained networks. It is worth noting about the time difference between these two pruning approaches. In case of pruning without retraining it is possible to prune the very deep networks from several minutes to 2-3 hours. The time depends on the size of the size of testing dataset. In case of using retraining many epochs should be run to achieve satisfactory level of sparsity with a very small drop in accuracy. In case of imagenet one epoch lasts about approximately one hour. The overall process takes few days. Choosing the method depends on hardware accelerator which will be used after pruning. If given hardware can make use of lower sparsity then pruning without retraining can be fast and efficient. In case of accelerator needs very high sparsity slow pruning with retraining should be performed. The last conclusion is that structure pruning without retraining doesn't guarantee low drop in accuracy. It should be run with retraining.

## 5 Further work

Further work will concentrate on tuning hyper-parameters in pruning algorithms which were described in a paper. It is still open problem if it is possible or how to find common rules for pruning all CNN networks to achieve satisfactory result. The next issue to focus on will be speeding up the pruning with retraining process by using more knowledge and statistics about the network. The proposed pruning methods of Deep Learning architectures can be also optimized and tested on a system level by taking data into consideration. This can be pronounced especially in latency critical systems [10].

Table 11: Results of structural pruning with retraining (CIFAR10)

| Name | baseline T1 | pruned T1 | pruned size |
|---|---|---|---|
| vgg19 | 92.37 | 92.42 | 52% |
| resnet50 | 95.26 | 94.98 | 72% |

Table 12: Results of structural pruning with retraining (CIFAR100)

| Name | baseline T1 | pruned T1 | pruned size |
|---|---|---|---|
| vgg19 | 70.62 | 70.71 | 54% |
| resnet50 | 78.21 | 77.50 | 78% |